RESEARCH ARTICLE                                                                                        OPEN ACCESS

# Removal of Salt and Pepper noise from Gray-Scale and Color Images: An Adaptive Approach

Sujaya Kumar Sathua [1], Arabinda Dash [2], Aishwaryarani Behera [3]
Department of CSE & IT [1], VSSUT, Burla
Department of CSE [2] & [3], Bhubaneswar College of Engineering, Khurda
Odisha - India

**ABSTRACT**
An efficient adaptive algorithm for the removal of Salt and Pepper noise from gray scale and color image is presented in this paper. In this proposed method first a 3X3 window is taken and the central pixel of the window is considered as the processing pixel. If the processing pixel is found as uncorrupted, then it is left unchanged. And if the processing pixel is found corrupted one, then the window size is increased according to the conditions given in the proposed algorithm. Finally the processing pixel or the central pixel is replaced by either the mean, median or trimmed value of the elements in the current window depending upon different conditions of the algorithm. The proposed algorithm efficiently removes noise at all densities with better Peak Signal to Noise Ratio (PSNR) and Image Enhancement Factor (IEF). The proposed algorithm is compared with different existing algorithms like MF, AMF, MDBUTMF, MDBPTGMF and AWMF.

## I. INTRODUCTION

An image may be defined as a two dimensional function, f(x, y) and it is formulated as

$$I = f(x, y) \qquad (1)$$

where x and y are spatial coordinates and I is the intensity or gray value at that point. When spatial coordinates and amplitude values are all finite, discrete quantities, then the image is called digital image. When a digital image is processed for receiving and analyzing visual information by digital computer, it is called as digital image processing [1]. A digital image is composed of a finite number of elements. These elements have a particular location and value, which is most widely known as pixel. The other terms used for the pixel are picture element, image element and pels. The digital image is represented by a single 2- dimensional integer array for a gray scale image and a series of three 2- dimensional arrays for each color bands.

Image restoration means to retrieve the clean image from the degraded version of the image by removing the unwanted noise. Noise present in the image can be of additive or multiplicative type depending upon how the image is formed. Impulse noise is one of the additive types of noise present in the image during signal acquisition stage or due to the bit error in the transmission. There are two types of impulse noise found in the image, they are random value impulse noise and fixed value impulse noise (which is known as Salt and Pepper noise). In salt and pepper noise the corrupted pixels take the maximum (i.e. 255) value or the minimum (i.e. 0) value which leads to white and black spots in the image. These noises in any form should be removed from the image before further processing. In this paper we have proposed an efficient adaptive algorithm for the removal of Salt and Pepper noise from the image.

Many algorithms have been proposed for the removal of salt and pepper noise from the image over the past two decades [2-9]. One of the most important issues in the image restoration is not only to remove noise but also to preserve the edge and texture details. To resolve this issue many good algorithms like Modified Decision Based Unsymmetric Median Filter (MDBUTMF) [7], Decision Based Partially Trimmed Global Mean Filter (DBPTGMF) [8], and Modified Decision Based Partially Trimmed Global Mean Filter (MDBPTGMF) [9] are proposed. In these algorithms, a fixed 3X3 window is taken and when a corrupted pixel is found then it is replaced by either the mean, median or trimmed value of the pixels inside the window. As the noise density increases these algorithms fails to preserve the texture details of the image i.e. the originality is lost at high noise density.

Due to the drawback of fixed window size, many adaptive schemes have been proposed for the removal of salt and pepper noise [10-15]. Among them Adaptive Weighted Mean Filter (AWMF) [15] which is proposed in the year 2014 performs well. In this method the window size is increased continuously until the maximum and minimum values of two successive windows are equal. If the given pixel value is equal to maximum or minimum values, then it is replaced by the weighted mean of the current window. Otherwise it is recognized as uncorrupted and remains unchanged. This algorithm has the lower detection error and better restoration image quality among all the adaptive methods proposed so far.

In this paper, we propose an efficient adaptive method for the removal of salt and pepper noise. The proposed algorithm is a variation to Modified Decision Based Partially Trimmed Global Mean Filter (MDBPTGMF) [9] as to overcome the drawback of fixed window size. In the proposed method, first a 3X3 window is taken where the central pixel is recognized as the processing pixel. Then if the processing pixel is found 0 or 255 then it is a noisy pixel, and the window size is increased depending upon the other elements of the current window. Lastly the processing pixel is replaced by either the mean, median or trimmed value of the elements in the current window according to the conditions given in the algorithm. The proposed algorithm is tested against various standard images and the experimental result shows that our proposed algorithm has better restoration image quality with better PSNR and IEF values as compare to other existing algorithms.

The outline of this paper is as follows. The MDBPTGMF filter is reviewed in section II. Our proposed algorithm is presented in section III. In section IV the illustration of proposed algorithm is given. Experimental and comparison result are given in section V. Finally we conclude our work in section VI.

## II. REVIEW OF MDBPTGMF

In Modified Decision Based Partially Trimmed Global Mean Filter (MDBPTGMF), first a 3X3 window is taken where the central pixel is recognised as the processing pixel. Then the processing pixel is checked to know whether it is noisy or noisefree. If it is uncorrupted then it is remain unchanged. And if the processing pixel is found to be noisy, then it is processed as per the algorithm given below:

*Algorithm of MDBPTGMF:*

Step 1: Select a 3X3 2-D window. Assume that the processing pixel is $P_{ij}$, which lies at the center of window.

Step 2: If $0 < P_{ij} < 255$, then the processing pixel or $P_{ij}$ is uncorrupted and left unchanged.

Step 3: If $P_{ij} = 0$ or $P_{ij} = 255$, then it is considered as corrupted pixel and four cases are possible as given below.

**Case i):** If the selected window has all the pixel value as 0, then replace $P_{ij}$ is by the Salt noise (i.e. 255).
**Case ii):** If the selected window contains all the pixel value as 255, then replace $P_{ij}$ by the pepper noise (i.e.0).
**Case iii):** If the selected window contains all the value as 0 and 255 both. Then replace processing pixel by the mean value of the current window.
**Case iv):** If the selected window contain not all the element 0 and 255. Then eliminate 0 and 255 and find the median value of the remaining element. Replace $P_{ij}$ with median value.

Step 4: Repeat step 1 to 3 for the entire image until the process is complete.

## III. PROPOSED ALGORITHM

To enhance the performance and overcome the drawback of fixed window size, based on the working mechanism of MDBPTGMF filter, the main aim behind the proposed algorithm is to preserve the edge and texture detail of the image.

In the proposed method, firstly a 3X3 2-D window is introduced. Then the central pixel is considered as the processing pixel. As we know in salt and pepper noise the corrupted pixel can take either the maximum (255) or the minimum (0) pixel value, then the processing pixel having 0 or 255 value is known as the corrupted one. And if the processing pixel value is between 0 and 255, then it is recognized as uncorrupted one and it left unchanged. Lastly if the processing pixel is found to be corrupted one then the size of the window and the restoration process is done according to the conditions in the algorithm given below:

Let, *W* is the current window size,

$W_{max}$ is the maximum window size,

$P_{ij}$ is the processing pixel,

*v* is the array of elements or pixels of current window *W*,

*N* is the number of uncorrupted pixel (pixel value lies in between 0 and 255) in the window *W*,

And *V* is the array of uncorrupted pixel of *W*.

*Algorithm:*

Step 1: Initialize $W = 3$, $h = 2$, $Wmax = 9$ and the central pixel of the window *W* is the processing pixel $P_{ij}$.

Step 2: If $0 < P_{ij} < 255$, then $P_{ij}$ is uncorrupted pixel and it left unchanged.

Step 3: If $P_{ij} = 0$ or $P_{ij} = 255$, then $P_{ij}$ is recognized as corrupted pixel and go to step-4.

Step 4: If $N \geq W$, then replace $P_{ij}$ with the median value of *V*. Else go to step-5.

Step 5: Process $P_{ij}$ according to the following cases given below

**Case i):** If $W < Wmax$ and $N < W$, then increment the window by $W = W + h$ up to Wmax.

**Case ii):** If $W < Wmax$ and $N \geq W$, then replace $P_{ij}$ with the median value of *V* and break the increment.

**Case iii):** If $W = Wmax$, $N < W$ and $N \neq 0$, then replace $P_{ij}$ with mean value of *V*.

**Case iv):** If $W = Wmax$ and *v* contain all the elements as 0 and 255, then replace $P_{ij}$ with the mean value of *v*.

**Case v):** If $W = Wmax$ and *v* contain all the elements as 0, then replace $P_{ij}$ with 255.

**Case vi):** If $W = Wmax$ and *v* contain all the elements as 255, then replace $P_{ij}$ with 0.

Step 6: Repeat step 2 to 5 for the entire image until the process is complete for all pixels.

## IV. ILLUSTRATION OF PROPOSED ALGORITHM

In the proposed algorithm, each and every pixel of the image is checked for the presence of salt and pepper noise by considering each and every pixel as the central pixel of the selected window. Let us discuss the proposed algorithm by taking an example. Considered a 9X9 matrix given below and our processing pixel is 255(underline pixel).

| 0 | 0 | 0 | 0 | 0 | 0 | 0 | 255 | 124 |
|---|---|---|---|---|---|---|---|---|
| 115 | 0 | 118 | 187 | 0 | 116 | 115 | 0 | 112 |
| 255 | 67 | 0 | 0 | 255 | 0 | 0 | 255 | 255 |
| 255 | 97 | 0 | 134 | 0 | 0 | 255 | 0 | 0 |
| 255 | 0 | 255 | 0 | **255** | 123 | 0 | 255 | 0 |
| 0 | 255 | 0 | 255 | 255 | 0 | 255 | 0 | 0 |
| 0 | 119 | 116 | 255 | 0 | 255 | 0 | 0 | 0 |
| 0 | 178 | 255 | 0 | 255 | 0 | 0 | 255 | 0 |
| 113 | 255 | 0 | 0 | 110 | 234 | 255 | 0 | 112 |

➢ From this 9X9 matrix, according to proposed algorithm first a 3X3 matrix is selected as given below:

| 134 | 0 | 0 |
|---|---|---|
| 0 | **255** | 123 |
| 255 | 255 | 0 |

➢ Here the central pixel is processing pixel, $P_{ij}$ = 255, so it is a noisy pixel.

➢ Here $W = 3$, $N = 2$ and $V = [123, 134]$, which means $W > N$. Therefore according to step 5: case i, the size of selected window is increased by 2. After increment the matrix will be as given below:

| 0 | 0 | 255 | 0 | 0 |
|---|---|---|---|---|
| 0 | 134 | 0 | 0 | 255 |
| 255 | 0 | **255** | 123 | 0 |
| 0 | 255 | 255 | 0 | 255 |
| 116 | 255 | 0 | 255 | 0 |

➢ Here $W = 5$, $N = 3$ and $V = [116, 123, 134]$, which means again $W > N$, as $W < W_{max}$ so again the window size will be incremented. The resultant matrix will be:

| 0 | 118 | 187 | 0 | 116 | 115 | 0 |
|---|---|---|---|---|---|---|
| 67 | 0 | 0 | 255 | 0 | 0 | 255 |
| 97 | 0 | 134 | 0 | 0 | 255 | 0 |
| 0 | 255 | 0 | **255** | 123 | 0 | 255 |
| 255 | 0 | 255 | 255 | 0 | 255 | 0 |
| 119 | 116 | 255 | 0 | 255 | 0 | 0 |
| 178 | 255 | 0 | 255 | 0 | 0 | 255 |

➢ Here $W = 7$, $N = 11$ and $V = [67, 97, 115, 116, 116, 118, 119, 123, 134, 178, 187]$, which means $W < N$. So according to condition step 5: case ii, of proposed algorithm, the central pixel is replaced by the median of $V$.

➢ So median ($V$) = 118, so 255 is replaced by 118. After this the resultant matrix will be:

| 134 | 0 | 0 |
|---|---|---|
| 0 | **118** | 123 |
| 255 | 255 | 0 |

## V. SIMULATION AND COMPARISON RESULTS

For simulation, we have taken 512 X 512 gray scale images like Lena and Goldhill images and color images i.e. rabbit and baboon. These images are corrupted with Salt and Pepper noise. The noise density is varied from 10% to 90%. These noises are removed by using proposed algorithm (PA) and different existing algorithms like Median Filter (MF), Adaptive Median Filter (AMF), Modified Decision Based Unsymmetric Trimmed Median Filter (MDBUTMF), Modified Decision Based Partially Trimmed Mean Filter (MDBPTGMF) and Adaptive Weighted Mean Filter (AWMF). We have executed the proposed algorithm and existing algorithms by 2.40 GHz, 2.00 GB RAM system and MATLAB 2009. We have measured the performance of proposed algorithm and existing algorithms by means of Peak Signal to Noise Ratio (PSNR) and Image Enhancement Factor (IEF). The performance of proposed algorithm is compared against different existing algorithms. MSE, PSNR and IEF are defined as

$$MSE = \frac{\sum_{i,j}(Y(i,j) - \hat{Y}(i,j))^2}{M \times N} \quad (2)$$

$$PSNR \text{ in dB} = 10 \log_{10}\left(\frac{255^2}{MSE}\right) \quad (3)$$

$$IEF = \frac{\sum_{i,j}(\eta(i,j) - Y(i,j))^2}{\sum_{i,j}(\hat{Y}(i,j) - Y(i,j))^2} \quad (4)$$

where, MSE stands for Mean Square Error, PSNR stands for Peak Signal to Noise Ratio, IEF stands for Image Enhancement Factor. M X N is the size of the image, Y represents original image. Ŷ represents restored image and η represents noisy image.

Different comparison results are shown in table I to VIII. From figure 1 to 8, we are showing the comparison between proposed algorithm and existing algorithms in terms of graph. In figures from 9 to 12, the original and noisy Lena, Goldhill, Baboon and Rabbit images and the restored images obtained by proposed algorithm and existing algorithms are shown.

In table I and II, the PSNR and IEF values of proposed algorithm (PA) and existing algorithms (like MF, AMF, MDBUTMF, MDBPTGMF and AWMF) for gray scale Lena image are shown respectively. The graphical comparison of PSNR and IEF values between the proposed algorithm and existing algorithm for gray scale Lena image are shown in figure 1 and 2.

TABLE-I
COMPARISION OF PSNR VALUES OF DIFFERENT ALGORITHMS FOR LENA IMAGE AT DIFFERENT NOISE DENSITIES

| Noise Density in % | PSNR in dB | | | | | |
|---|---|---|---|---|---|---|
| | MF | AMF | MDBUT MF | MDBPT GMF | AWMF | PA |
| 10 | 26.84 | 36.91 | 39.98 | 39.01 | 41.98 | 45.12 |
| 20 | 24.91 | 34.92 | 37.21 | 37.11 | 39.30 | 42.27 |
| 30 | 24.80 | 32.98 | 36.01 | 36.52 | 38.06 | 40.76 |
| 40 | 24.65 | 31.62 | 35.13 | 35.71 | 36.98 | 38.88 |
| 50 | 24.37 | 29.11 | 33.71 | 34.79 | 35.09 | 36.23 |
| 60 | 23.97 | 27.44 | 31.54 | 32.75 | 33.97 | 34.57 |
| 70 | 23.10 | 26.01 | 28.81 | 29.13 | 30.61 | 32.99 |
| 80 | 20.01 | 23.82 | 26.18 | 27.83 | 29.01 | 31.03 |
| 90 | 11.26 | 21.36 | 24.06 | 25.18 | 26.13 | 29.61 |

TABLE-II
COMPARISION OF IEF VALUES OF DIFFERENT ALGORITHMS FOR LENA IMAGE AT DIFFERENT NOISE DENSITIES

| Noise Density in % | PSNR in dB | | | | | |
|---|---|---|---|---|---|---|
| | MF | AMF | MDBUT MF | MDBPT GMF | AWMF | PA |
| 10 | 10.3 | 240.6 | 594.9 | 531.1 | 295.5 | 727.6 |
| 20 | 28.1 | 159.6 | 444.5 | 415.7 | 263.9 | 663.9 |
| 30 | 30.1 | 114.7 | 465.1 | 398.6 | 242.8 | 546.6 |
| 40 | 23.1 | 85.9 | 323.5 | 343.3 | 131.4 | 456.1 |
| 50 | 11.7 | 72.3 | 287.5 | 292.4 | 102.9 | 386.7 |
| 60 | 6.7 | 44.59 | 170.5 | 197.9 | 52.1 | 301.1 |
| 70 | 3.3 | 18.3 | 98.6 | 115.9 | 24.2 | 223.8 |
| 80 | 2.0 | 5.4 | 56.7 | 84.7 | 11.2 | 170.4 |
| 90 | 1.4 | 3.2 | 12.9 | 18.1 | 6.1 | 128.1 |

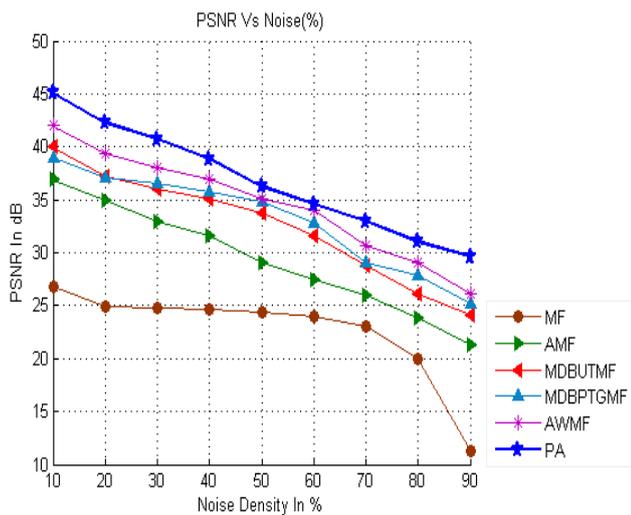

Fig. 1.Comparison graph of PSNR at different noise densities for 'Lena' Image

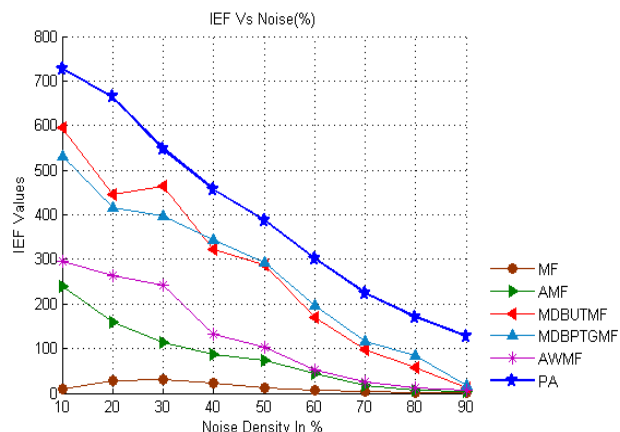

Fig. 2.Comparison graph of IEF at different noise densities for

From table I & II and figure 1 & 2, it can be clearly seen that the performance of proposed algorithm is best among all the existing algorithms in terms of PSNR and IEF.

In order to present some extra evidence that our method has the best result than other existing methods, in table III & IV and figure 3 & 4 the tabulation and graphical comparison of PSNR and IEF values of gray scale Goldhill image are given respectively. In figure 5 and 6 the restored images obtained by proposed algorithm and existing algorithms are shown for gray scale Lena and Goldhill images respectively.

TABLE-III
COMPARISION OF PSNR VALUES OF DIFFERENT ALGORITHMS FOR GOLDHILL IMAGE AT DIFFERENT NOISE DENSITIES

| Noise Density in % | PSNR in dB | | | | | |
|---|---|---|---|---|---|---|
| | MF | AMF | MDBUT MF | MDBPT GMF | AWMF | PA |
| 10 | 24.21 | 33.14 | 36.31 | 34.28 | 35.21 | 41.96 |
| 20 | 23.46 | 31.93 | 34.92 | 33.75 | 33.47 | 37.58 |
| 30 | 23.25 | 29.20 | 33.64 | 32.35 | 31.94 | 35.36 |
| 40 | 23.05 | 27.11 | 32.97 | 32.37 | 30.62 | 33.68 |
| 50 | 22.72 | 26.08 | 32.13 | 31.68 | 29.34 | 32.13 |
| 60 | 22.51 | 25.17 | 30.75 | 30.47 | 28.14 | 30.72 |
| 70 | 20.44 | 23.91 | 28.18 | 29.01 | 26.48 | 29.47 |
| 80 | 18.22 | 22.64 | 26.74 | 27.64 | 25.69 | 27.83 |
| 90 | 12.79 | 20.01 | 24.94 | 25.91 | 23.96 | 25.92 |

TABLE-IV
COMPARISION OF IEF VALUES OF DIFFERENT ALGORITHMS FOR GOLDHILL IMAGE AT DIFFERENT NOISE DENSITIES

| Noise Density in % | PSNR in dB | | | | | |
|---|---|---|---|---|---|---|
| | MF | AMF | MDBUT MF | MDBPT GMF | AWMF | PA |
| 10 | 32.3 | 120.7 | 156.1 | 126.5 | 157.8 | 389.3 |
| 20 | 33.1 | 90.1 | 172.7 | 148.9 | 145.9 | 265.5 |
| 30 | 17.5 | 73.1 | 184.5 | 165.4 | 132.5 | 239.2 |
| 40 | 8.6 | 59.9 | 78.8 | 66.4 | 74.1 | 215.0 |
| 50 | 4.6 | 49.0 | 52.0 | 44.8 | 58.8 | 187.8 |
| 60 | 2.9 | 35.9 | 23.1 | 21.8 | 40.9 | 164.0 |
| 70 | 2.0 | 16.1 | 10.1 | 9.7 | 21.2 | 143.3 |
| 80 | 1.5 | 5.2 | 4.6 | 4.6 | 8.9 | 112.1 |
| 90 | 1.2 | 2.0 | 2.2 | 3.3 | 7.6 | 83.2 |

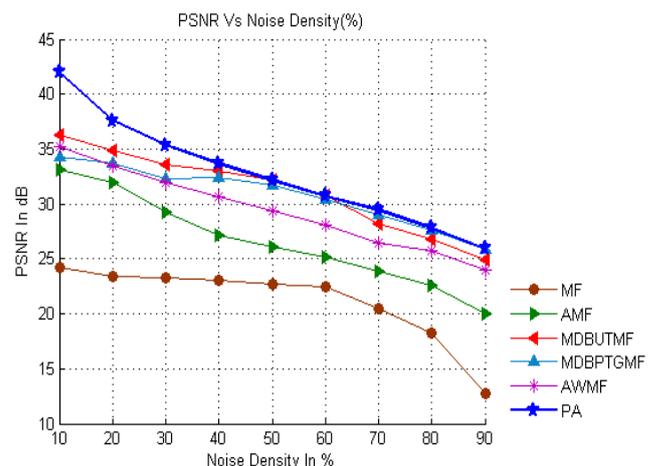

Fig. 3.Comparison graph of PSNR at different noise densities for 'Goldhill' Image

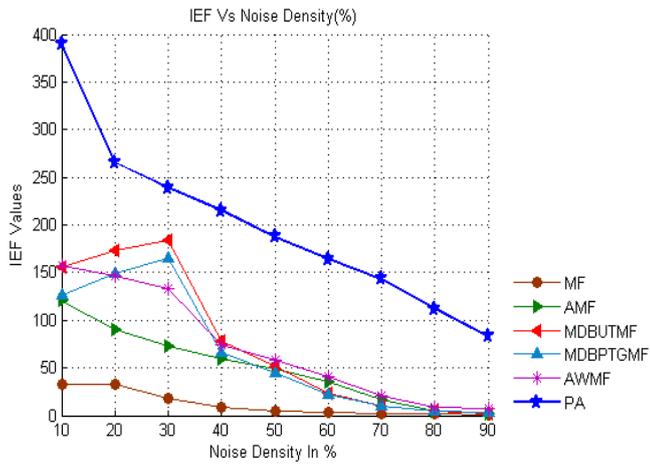

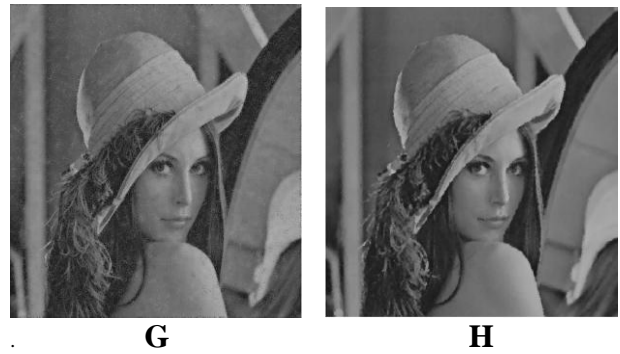

G      H

Fig.5.Performance of various algorithms for gray scale Lena image. (A) Original image. (B) Corrupted image with 70% salt and pepper noise. (C) MF. (D) AMF. (E) MDBUTMF. (F) MDBPTGMF. (G) AWMF. (H) PA

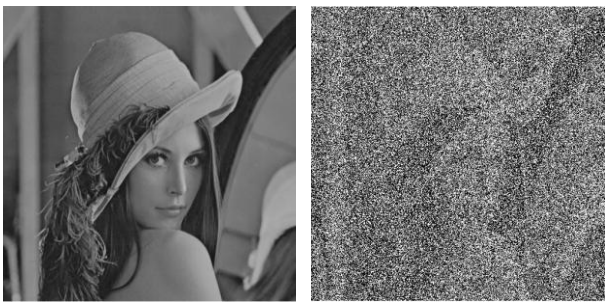

A      B

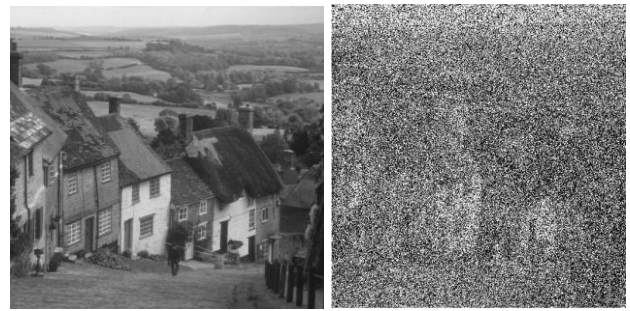

A      B

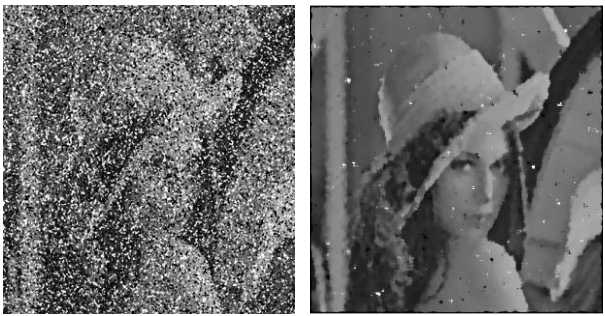

C      D

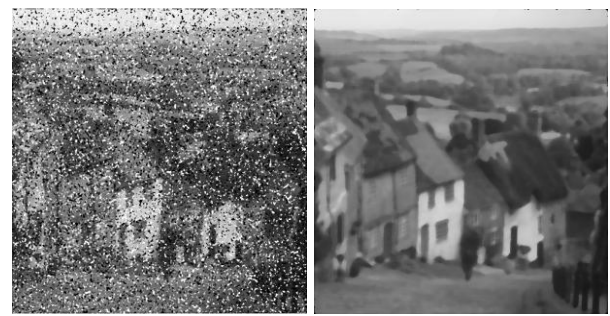

C      D

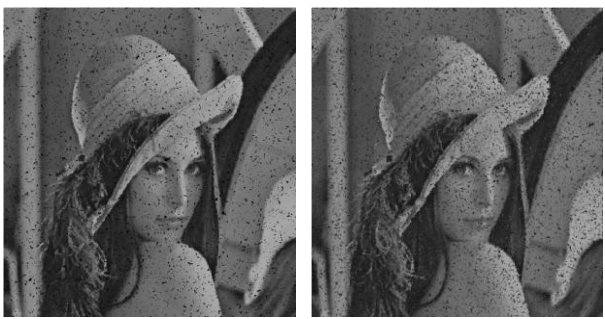

E      F

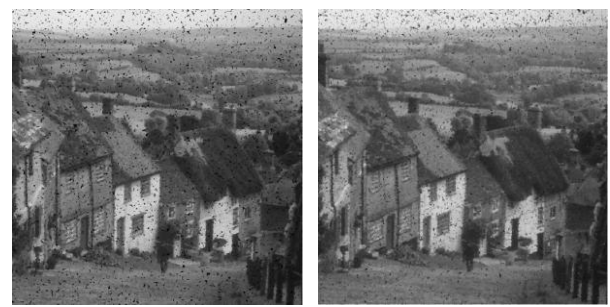

E      F

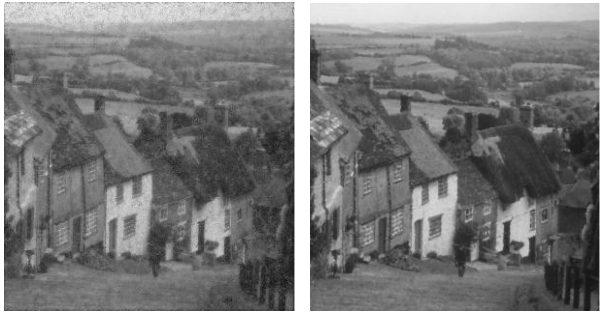

**G**          **H**

Fig.5.Performance of various algorithms for gray scale Lena image. (A) Original image. (B) Corrupted image with 70% salt and pepper noise. (C) MF. (D) AMF. (E) MDBUTMF. (F) MDBPTGMF. (G) AWMF. (H) PA

From figure 5 and 6, it can be clearly seen than our proposed method is best among all the existing method for gray scale images. The above comparisons are given for gray scale images.

Now we are presenting the performance of our proposed method for color images like Baboon and Rabbit. In table V and VI the comparison of PSNR and IEF for color Baboon image is given. Similarly in table VII and VIII the PSNR and IEF values for color Rabbit image are shown respectively.

TABLE-V
COMPARISION OF PSNR VALUES OF DIFFERENT ALGORITHMS FOR BABOON IMAGE AT DIFFERENT NOISE DENSITIES

| Noise Density in % | PSNR in dB | | | | | |
|---|---|---|---|---|---|---|
| | MF | AMF | MDBUT MF | MDBPT GMF | AWMF | PA |
| 10 | 22.70 | 30.01 | 25.99 | 21.81 | 31.18 | 33.62 |
| 20 | 21.64 | 26.85 | 23.44 | 21.69 | 28.06 | 30.66 |
| 30 | 19.70 | 24.61 | 22.50 | 21.46 | 25.52 | 28.71 |
| 40 | 16.92 | 22.70 | 21.76 | 20.99 | 22.46 | 27.18 |
| 50 | 14.27 | 21.01 | 20.39 | 19.88 | 21.78 | 26.04 |
| 60 | 11.85 | 19.39 | 18.23 | 17.69 | 20.17 | 24.99 |
| 70 | 9.69 | 16.66 | 15.05 | 14.80 | 18.72 | 23.17 |
| 80 | 7.92 | 12.45 | 11.68 | 11.39 | 16.45 | 22.22 |
| 90 | 6.52 | 8.53 | 8.36 | 8.33 | 14.30 | 21.26 |

TABLE-VI
COMPARISION OF IEF VALUES OF DIFFERENT ALGORITHMS FOR BABOON IMAGE AT DIFFERENT NOISE DENSITIES

| Noise Density in % | PSNR in dB | | | | | |
|---|---|---|---|---|---|---|
| | MF | AMF | MDBUT MF | MDBPT GMF | AWMF | PA |
| 10 | 5.4 | 30.2 | 11.5 | 4.3 | 34.5 | 42.5 |
| 20 | 8.5 | 28.2 | 23.2 | 8.7 | 28.7 | 42.9 |
| 30 | 8.1 | 25.7 | 15.6 | 12.4 | 25.7 | 41.4 |
| 40 | 5.7 | 21.9 | 17.6 | 14.6 | 22.4 | 38.7 |
| 50 | 3.9 | 18.9 | 16.1 | 14.2 | 18.9 | 36.9 |
| 60 | 2.7 | 15.3 | 11.7 | 10.1 | 16.5 | 35.0 |
| 70 | 1.9 | 9.5 | 6.5 | 6.1 | 10.6 | 33.9 |
| 80 | 1.4 | 4.1 | 3.4 | 3.2 | 8.2 | 30.9 |
| 90 | 1.1 | 1.9 | 1.8 | 1.8 | 7.0 | 27.9 |

TABLE-VII
COMPARISION OF PSNR VALUES OF DIFFERENT ALGORITHMS FOR RABBIT IMAGE AT DIFFERENT NOISE DENSITIES

| Noise Density in % | PSNR in dB | | | | | |
|---|---|---|---|---|---|---|
| | MF | AMF | MDBUT MF | MDBPT GMF | AWMF | PA |
| 10 | 32.69 | 32.58 | 29.28 | 29.05 | 34.21 | 41.19 |
| 20 | 28.31 | 28.86 | 28.17 | 27.91 | 30.35 | 38.35 |
| 30 | 23.01 | 26.51 | 27.06 | 26.62 | 26.65 | 36.34 |
| 40 | 18.20 | 24.62 | 25.11 | 24.74 | 25.73 | 34.58 |
| 50 | 14.55 | 22.80 | 22.15 | 21.63 | 22.25 | 33.10 |
| 60 | 11.57 | 20.83 | 18.23 | 17.71 | 20.22 | 31.48 |
| 70 | 9.25 | 15.84 | 14.02 | 13.75 | 17.44 | 29.98 |
| 80 | 7.41 | 12.68 | 10.21 | 10.02 | 14.85 | 28.22 |
| 90 | 5.88 | 8.07 | 6.77 | 6.70 | 13.45 | 26.02 |

TABLE-VIII
COMPARISION OF IEF VALUES OF DIFFERENT ALGORITHMS FOR RABBIT IMAGE AT DIFFERENT NOISE DENSITIES

| Noise Density in % | PSNR in dB | | | | | |
|---|---|---|---|---|---|---|
| | MF | AMF | MDBUT MF | MDBPT GMF | AWMF | PA |
| 10 | 63.5 | 78.0 | 29.0 | 27.2 | 156.5 | 448.9 |
| 20 | 46.1 | 66.4 | 44.4 | 42.0 | 116.8 | 466.8 |
| 30 | 20.4 | 57.5 | 52.1 | 46.9 | 83.6 | 439.8 |
| 40 | 9.0 | 49.7 | 44.2 | 40.5 | 46.1 | 391.4 |
| 50 | 4.8 | 40.7 | 28.1 | 24.7 | 31.3 | 347.9 |
| 60 | 2.93 | 31.0 | 13.5 | 12.0 | 28.5 | 287.5 |
| 70 | 2.0 | 14.5 | 6.0 | 5.6 | 16.6 | 237.6 |
| 80 | 1.5 | 5.0 | 2.7 | 2.8 | 5.2 | 181.1 |
| 90 | 1.2 | 2.0 | 1.4 | 1.4 | 4.3 | 122.5 |

In figure from 7 to 10, we are showing the graphical representation of comparison results shown in above tables for color Baboon and Rabbit images. From these graphical representation it can be seen that the proposed algorithm shown the best result for color images too as compare to existing algorithms.

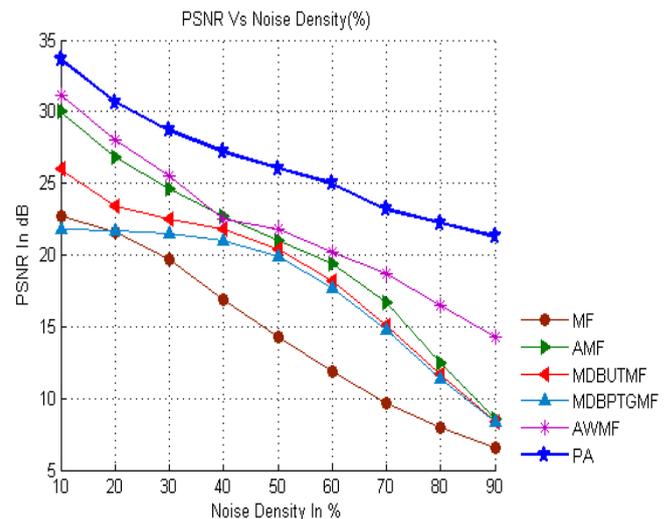

Fig. 7.Comparison graph of PSNR at different noise densities for 'Baboon' Image

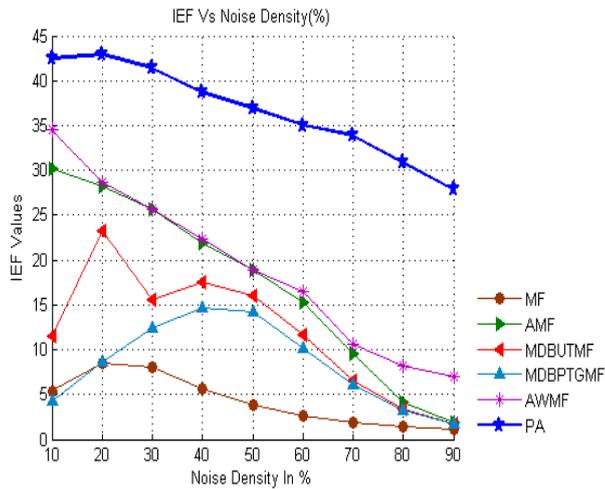

Fig. 8.Comparison graph of IEF at different noise densities for 'Baboon' Image

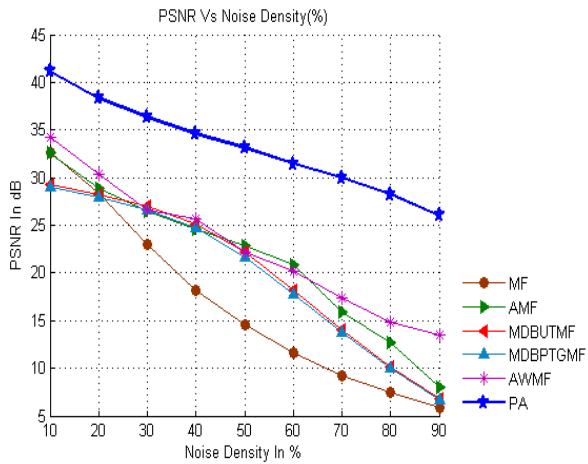

Fig. 9.Comparison graph of PSNR at different noise densities for 'Rabbit' Image

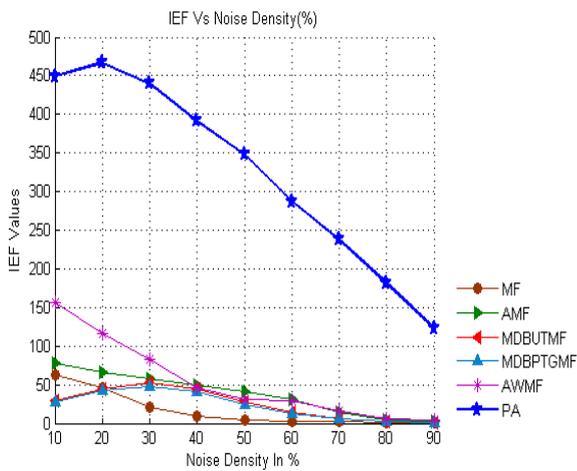

Fig. 10.Comparison graph of IEF at different noise densities for 'Rabbit' Image

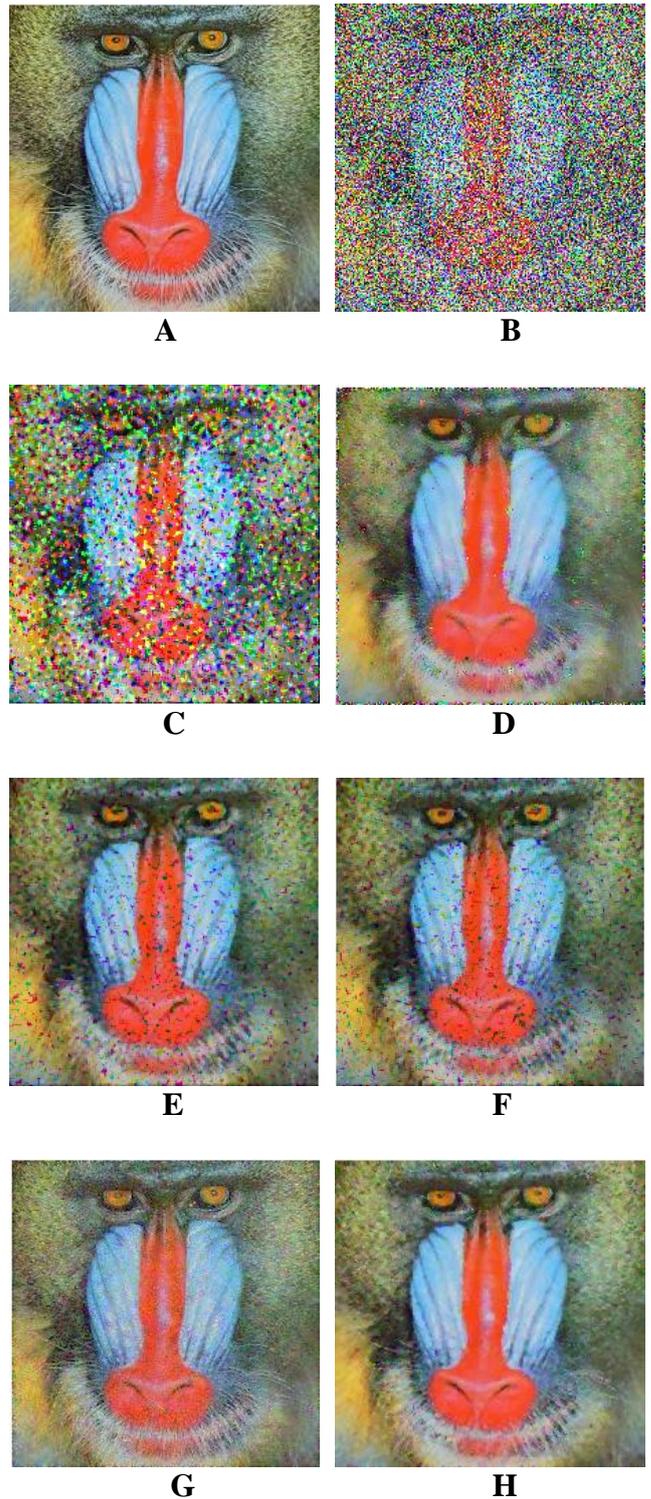

Fig.11.Performance of various algorithms for color Baboon image. (A) Original image. (B) Corrupted image with 60% salt and pepper noise. (C) MF. (D) AMF. (E) MDBUTMF. (F) MDBPTGMF. (G) AWMF. (H) PA.

Now in figure 11 and 12, we are showing the restored images by proposed algorithm and existing algorithms for color images like Baboon and Rabbit which are corrupted by 60% and 70% of Salt and Pepper noise respectively. Figure 11 and 12 clearly showing that the proposed algorithm is showing better results.

Finally we are presenting some extra evidence of the denoising performance of proposed method for gray scale as well as color images with different image formats like jpg, gif, bmp and tiff. In figure 13 and 14, the restored gray scale Cameraman.tif and Child.png obtained by the proposed algorithm are shown. Similarly, the restored color images i.e. Finger.jpg and Dog.bmp obtained by the proposed algorithm are shown in figure 15 and 16 respectively. From these figures it can be clearly seen that the proposed method have the better denoising capability for gray scale and color images.

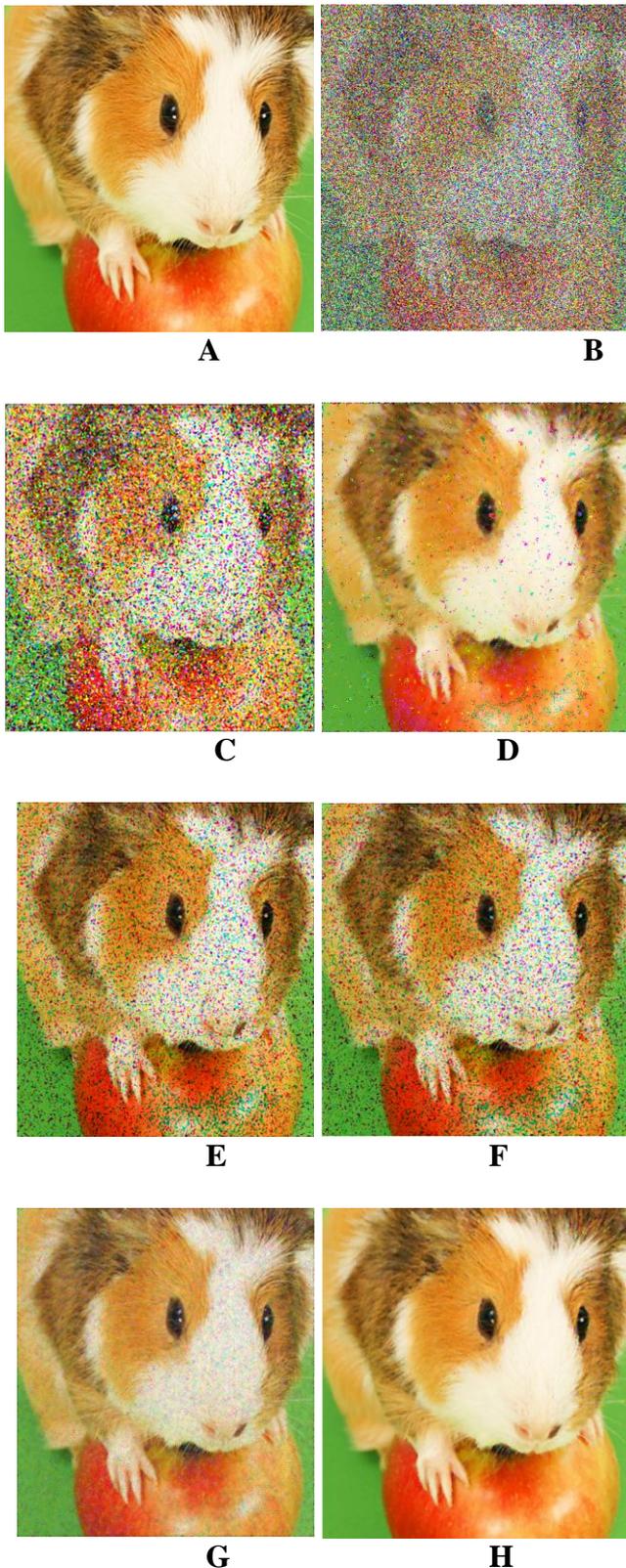

Fig.12.Performance of various algorithms for color Rabbit image. (A) Original image. (B) Corrupted image with 70% salt and pepper noise. (C) MF. (D) AMF. (E) MDBUTMF. (F) MDBPTGMF. (G) AWMF. (H) PA.

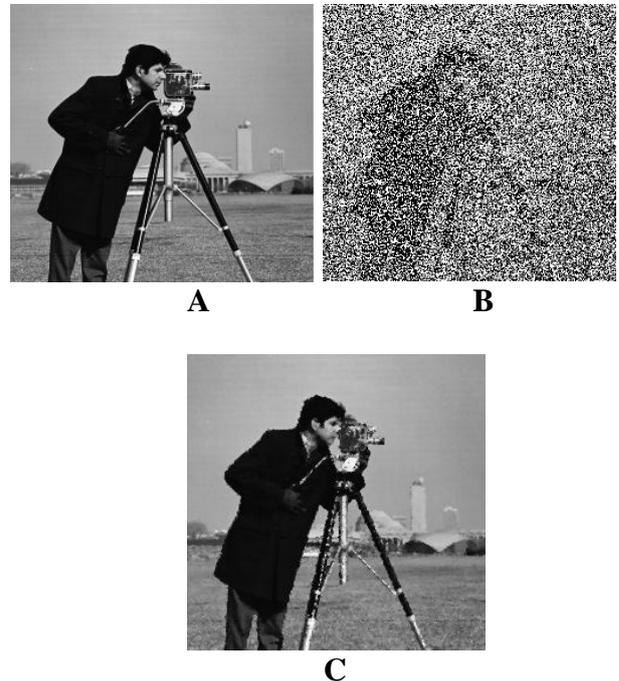

Fig.13.Performance of proposed algorithm for gray scale Cameraman.tif image. (A) Original image. (B) Corrupted image with 70% salt and pepper noise. (C) Restored Image with PSNR 23.13 dB.

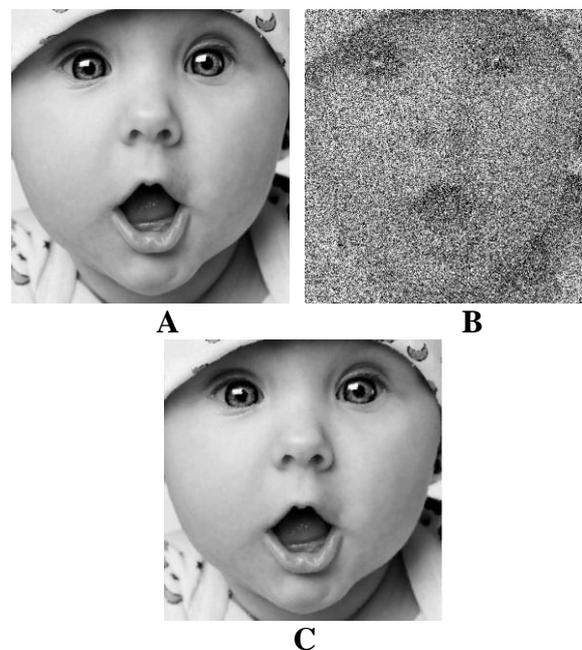

Fig.14.Performance of proposed algorithm for gray scale Child.png image. (A) Original image. (B) Corrupted image with 70% salt and pepper noise.(C) Restored Image with PSNR 36.14 dB.

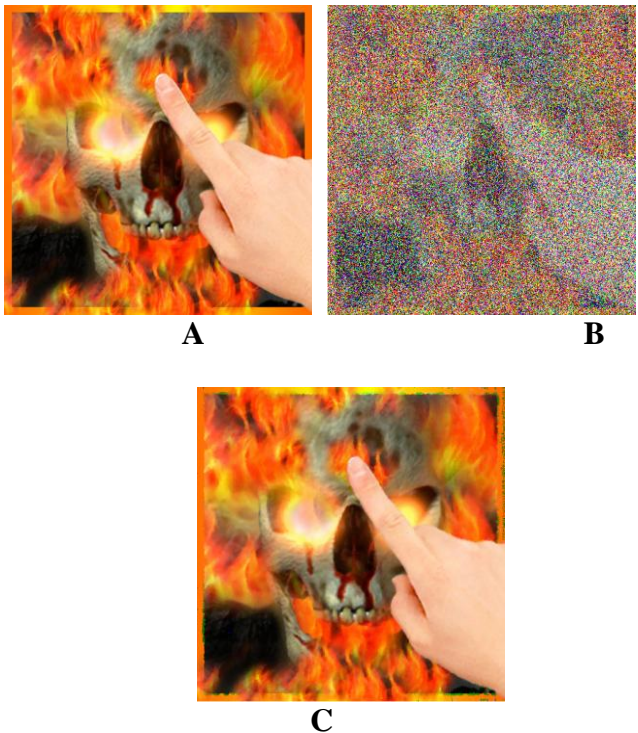

Fig.15.Performance of proposed algorithm for color Finger.jpg image. (A) Original image. (B) Corrupted image with 70% salt and pepper noise. (C) Restored Image.

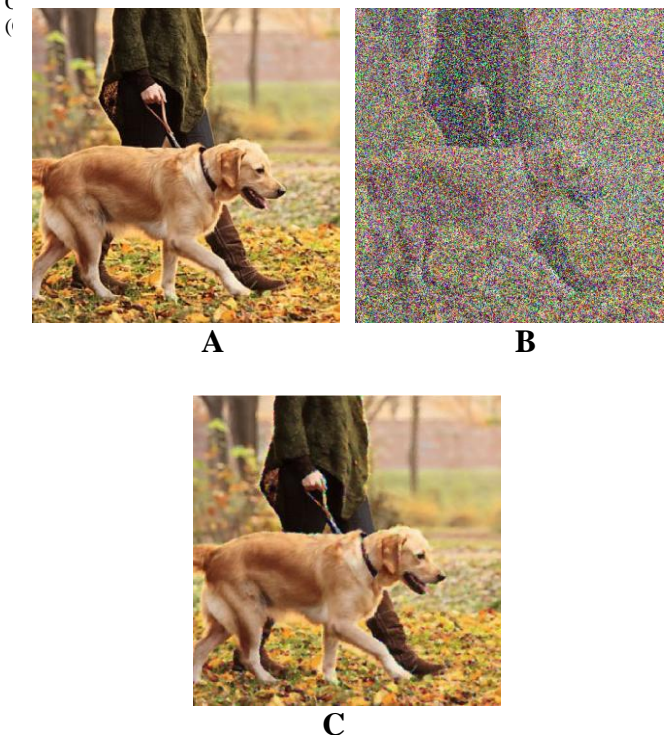

Fig.16.Performance of proposed algorithm for color Dog.bmp image. (A) Original image. (B) Corrupted image with 70% salt and pepper noise. (C) Restored Image with PSNR 24.79 dB.

## VI. CONCLUSION

In this paper, we have proposed an efficient adaptive method based on MDBPTGMF algorithm which perform better in restoring gray scale as well as color images corrupted by Salt and Pepper noise. The experimental results show that our proposed adaptive algorithm gives better result in terms of PSNR and IEF values as compare to other existing algorithms. As in the proposed method, the noise is detected by comparing the pixels of image directly with 0 or 255 value; therefore it has no detection error. The proposed method works for gray scale images and color images as well as it perform well for all image formats.